\documentclass{article}
\usepackage{spconf,amsmath,graphicx, booktabs, multirow, makecell}

\usepackage{enumitem}
\setlist{nosep, leftmargin=14pt}

\usepackage{float}
\usepackage{subcaption}
\usepackage{url}
\title{Benchmarking Deep Learning for Future Liver Remnant Segmentation in Colorectal Liver Metastasis}
\name{\begin{tabular}{c}
    Anthony Wu$^{1,2*}$, Arghavan Rezvani$^{1*}$, Kela Liu$^{1*}$, Roozbeh Houshyar$^{2}$, Pooya Khosravi$^{1,2}$ \\
    Whitney Li$^{2}$ and Xiaohui Xie$^{1}$
      \end{tabular}
}
\address{$^{1}$ University of California, Irvine Donald Bren School of Information and Computer Science \\{$^{2}$} University of California, Irvine School of Medicine}
\begin{document}
%
\maketitle

\section{Abstract}
Accurate segmentation of the future liver remnant (FLR) is critical for surgical planning in colorectal liver metastases (CRLM) to prevent fatal post-hepatectomy liver failure. However, this segmentation task is technically challenging due to complex resection boundaries, convoluted hepatic vasculature and diffuse metastatic lesions. A primary bottleneck in developing automated AI tools has been the lack of high-fidelity, validated data. We address this gap by manually refining all 197 volumes from the public CRLM-CT-Seg dataset, creating the first open-source, validated benchmark for this task. We then establish the first segmentation baselines, comparing cascaded (Liver \(\rightarrow\) CRLM \(\rightarrow\) FLR) and end-to-end (E2E) strategies using nnU-Net, SwinUNETR, and STU-Net. We find a cascaded nnU-Net achieves the best final FLR segmentation Dice ($0.767$), while the pretrained STU-Net provides superior CRLM segmentation ($0.620$ Dice) and is significantly more robust to cascaded errors. This work provides the first validated benchmark and a reproducible framework to accelerate research in AI-assisted surgical planning.

\section{Introduction}

Colorectal cancer (CRC) is the second leading cause of cancer-related mortality, with nearly two million new cases reported annually~\cite{bray2024global}. Colorectal liver metastases (CRLM) is a major driver of this mortality, affecting  $\sim50\%$ of CRC patients \cite{kron2024new, rompianesi2022artificial} and is the principal cause of death in nearly half of those cases \cite{helling2014cause}. Surgical liver resection, the only curative option for CRLM (5-year survival $\sim$40\%~\cite{chow2019colorectal, calderon2023pushing}), is highly challenging. Planners must navigate complex anatomy and subtle lesions (``vanishing'' metastases)~\cite{veerankutty2021artificial, salavracos2024contribution} to preserve a sufficient future liver remnant (FLR) and prevent post-hepatectomy liver failure (PHLF), a major cause of mortality~\cite{kauffmann2014post}.

Previous studies have applied artificial intelligence (AI) techniques to various CRLM-related tasks such as identifying metastatic lesions~\cite{rompianesi2022artificial}, predicting recurrence~\cite{rompianesi2022artificial}, or assessing surgical candidacy~\cite{amygdalos2023novel, mehrabi2018meta}. Yet, few efforts directly address \textbf{surgical resection planning (SRP)}---that is, identifying which hepatic regions to resect and to preserve. 

A major hindrance to developing such a tool is data scarcity. Creating SRP datasets requires paired pre-operative imaging and detailed post-operative delineations—a labor-intensive and clinically specialized task. Defining these resection boundaries is inherently subjective, suffering from high inter-rater variability even among expert surgeons. Furthermore, segmenting CRLM is technically difficult, with `vanishing' metastases, small satellite foci, and partial volume effects making a `ground truth' notoriously hard to establish. The recently introduced \emph{CRLM-CT-Seg} dataset~\cite{CRLMData} represents a major advance, providing 197 CT scans with expert-generated semi-automatic segmentations of the liver, metastases, and FLR. While invaluable for organ and tumor segmentation studies, its semi-automatic nature introduced artifacts and irregular boundaries that have limited its direct use for high-fidelity SRP.

In this work, we build on this foundational \emph{CRLM-CT-Seg} dataset by manually refining annotations for all 197 CT volumes to unlock their full potential for SRP. Our \textbf{contributions} are threefold: (1) a publicly available, fully validated extension of CRLM-CT-Seg designed for surgical resection modeling\footnote{The refined dataset will be made publicly available on Zenodo upon publication at \url{https://doi.org/10.5281/zenodo.17574862}}; (2) the first benchmark across multiple architectures for FLR prediction; and (3) a reproducible framework to support future research in AI-assisted surgical planning.

\begin{figure*}[htbp] 
\centering
\includegraphics[width=1\linewidth]{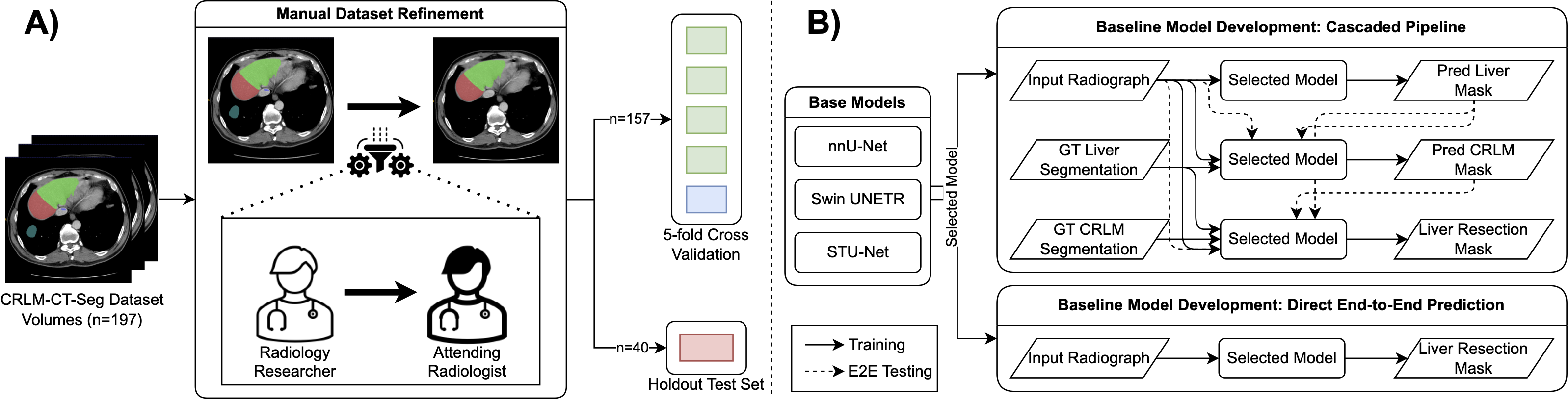} 
\caption{
\textbf{Study Overview.} \textbf{(a)} Dataset refinement (radiologist-confirmed) and data split. An example correction of an erroneous CRLM label in the lung (blue) is shown. Labels: LRS (red), FLR (green). \textbf{(b)} Model development comparing a 3-stage Cascaded pipeline (top) and an End-to-End (E2E) strategy (bottom). For the cascade, solid lines = training (ground-truth inputs); dotted lines = inference (predicted inputs).
}
\label{fig:main_figure}
\vspace{-10pt}
\end{figure*}


\begin{figure}[htb]
\centering
 \includegraphics[width=.95\linewidth]{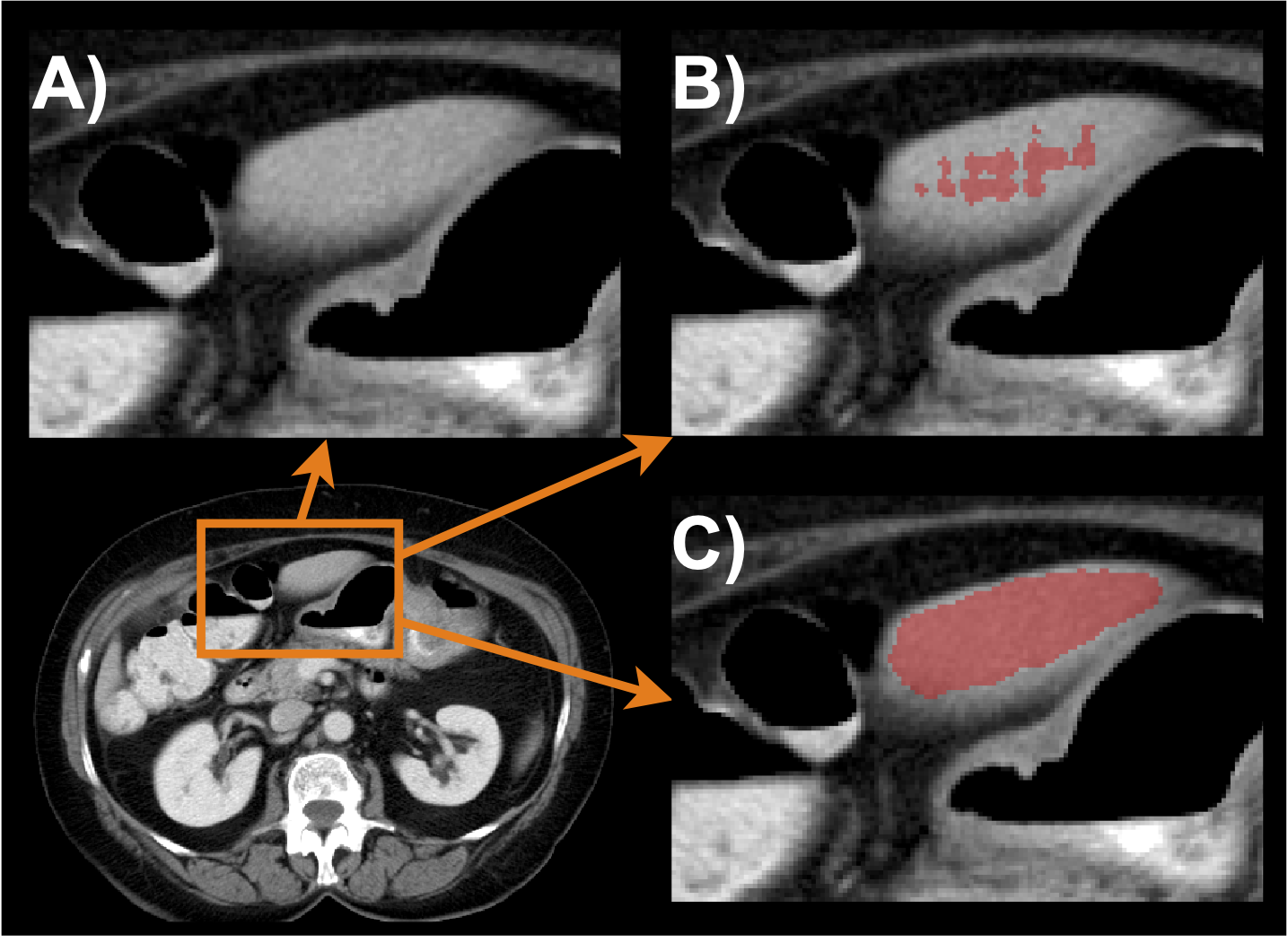} \caption{Refinement of liver border. (A) Axial CT (W/L 370/70 HU) shows two lesions (arrows). (B) Original segmentation contains patch-like artifacts. (C) Refined manual segmentation removes voids and ensures precise liver contours.}
\label{fig:semiAutoRemnantFix}
\vspace{-15pt}
\end{figure}






\section{Materials and Methods}

\textbf{3.1 Dataset and Manual Refinement:} We manually refined all 197 volumetric segmentations from the CRLM-CT-Seg dataset~\cite{CRLMData} using ITK-Snap~\cite{itksnap}. All refinements were performed by a radiological researcher with final refinements confirmed by an abdominal radiologist with over ten years of subspecialty experience (Fig.~\ref{fig:main_figure}a). Refinement focused on (1) liver, (2) FLR, (3) Liver Resection Segmentation ($Liver - FLR$), and (4) CRLM. The original segmentations exhibit edge irregularities for liver and FLR masks, likely arising from the semi-automated segmentation approach described in the original publication \cite{CRLMData}. These were corrected to produce smooth, anatomically consistent parenchymal boundaries (Fig.~\ref{fig:semiAutoRemnantFix}). 

\begin{figure}[htb]
\centering
 \includegraphics[width=0.9\linewidth]{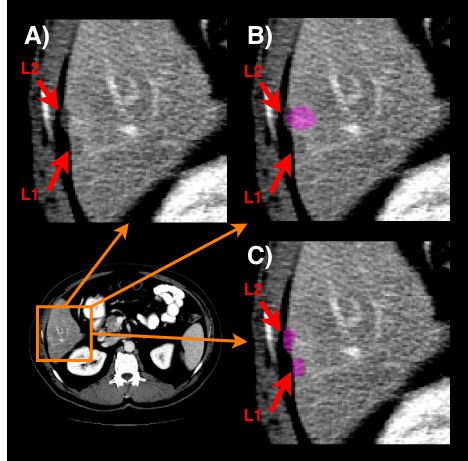} \caption{Refinement of imprecise CRLM segmentation. (A) Axial CT (W/L 178/88 HU) shows two lesions (arrows). (B) Original segmentation fails to capture lesion L1 and coarsely over-segments L2. (C) Refined manual segmentation provides precise contours for both.}
\label{fig:tumorSegFix}
\vspace{-10pt}
\end{figure}

For CRLM, lesion contours were manually reviewed to ensure accurate depiction of extent. Very small punctate foci with uncertain classification (i.e., possible satellite lesions vs. partial volume artifact) were intentionally excluded to maintain label specificity (Fig.~\ref{fig:tumorSegFix}).

\noindent
\textbf{3.2 Baseline Strategies for FLR Prediction:} The dataset was divided into 157 training-validation cases and 40 held-out test cases ($\sim$4:1 ratio). We compare two strategies for FLR prediction: a direct End-to-end (E2E) approach and a three-stage cascade. The former receives the CT volume alone and directly predicts FLR. The latter decomposes the task into three independently trained segmentation models:\\

\qquad $\textbf{Stage 1:} \quad \{\text{CT}\} \rightarrow \text{Liver}$

\qquad $\textbf{Stage 2:} \quad \{\text{CT},\ \text{Liver}\} \rightarrow \text{CRLM}$

\qquad $\textbf{Stage 3:} \quad \{\text{CT},\ \text{Liver},\ \text{CRLM}\} \rightarrow \text{FLR}$

In Stages 2 and 3, the liver mask is used to mask the CT input (i.e., cropping to the liver ROI). In Stage 3, the CRLM mask is concatenated as an additional input channel. During training, each stage uses ground-truth inputs (e.g., the CRLM model receives the ground truth liver mask). During inference, the stages take the predicted outputs generated by the previous stage (Fig. ~\ref{fig:main_figure}b). For both strategies, we evaluate three representative 3D segmentation architectures: nnU-Net~\cite{nnUnet}, SwinUNETR~\cite{swinUnetr}, and STU-Net~\cite{stunet}.

\noindent
\textbf{3.3 Implementation and Evaluation:} Base model sizes were used. nnU-Net and SwinUNETR were trained from scratch; STU-Net was initialized using official pretrained weights as per author recommendation. All models trained for 200 epochs using AdamW, binary cross entropy loss, and nnU-Net’s standard medical imaging augmentations with recommended hyperparameters per each models' authors.

All models were trained on the 157-case training-validation subset using a 5-fold cross-validation (CV) scheme. To demonstrate training stability, we report mean $\pm$ std CV metrics. Final performance for all strategies is reported on the 40-case held-out test set by ensembling the logits from the five trained folds, consistent with standard nnU-Net inference. We evaluate all individual cascade stages, the final cascaded E2E FLR, and the direct E2E FLR using Dice, sensitivity (recall), and precision.

\begin{table}[t]
\centering
\caption{Model 5-fold cross-validation performance. All metrics are macro-averaged on a per-case basis. Abbreviations: D (Dice), P (Precision), R (Recall), L (Liver), T (CRLM Tumor), E2E (End-to-End).}
\label{tab:cross_val_results}
\setlength{\tabcolsep}{.6mm}
\resizebox{\columnwidth}{!}{
\begin{tabular}{l c cccc}
\toprule
 & & \multicolumn{3}{c}{\textbf{Pipelined Tasks}} & \multicolumn{1}{c}{\textbf{E2E Task}} \\
\cmidrule(r){3-5} \cmidrule(l){6-6}
\textbf{Model} & & 
\makecell{\textbf{Liver} \\ \textbf{(CT)}} & 
\makecell{\textbf{CRLM} \\ \textbf{(L+CT)}} & 
\makecell{\textbf{FLR} \\ \textbf{(L+T+CT)}} &
\makecell{\textbf{FLR} \\ \textbf{(CT)}} \\
\midrule
\multirow{3}{*}{\textbf{nnU-Net}} 
 & D & $0.951 \pm 0.009$ & $0.689 \pm 0.041$ & $\mathbf{0.834 \pm 0.023}$ & $0.755 \pm 0.027$ \\
 & P & $0.940 \pm 0.012$ & $0.740 \pm 0.057$ & $\mathbf{0.822 \pm 0.025}$ & $0.677 \pm 0.042$ \\
 & R & $0.965 \pm 0.008$ & $0.703 \pm 0.053$ & $0.903 \pm 0.029$ & $\mathbf{0.923 \pm 0.010}$ \\
\midrule
\multirow{3}{*}{\makecell{\textbf{Swin-} \\ \textbf{UNETR}}} 
 & D & $0.960 \pm 0.012$ & $0.573 \pm 0.055$ & $0.756 \pm 0.037$ & $0.737 \pm 0.032$ \\
 & P & $0.959 \pm 0.017$ & $\mathbf{0.784 \pm 0.051}$ & $0.732 \pm 0.042$ & $\mathbf{0.717 \pm 0.053}$ \\
 & R & $0.961 \pm 0.009$ & $0.529 \pm 0.071$ & $0.873 \pm 0.042$ & $0.829 \pm 0.034$ \\
\midrule
\multirow{3}{*}{\textbf{STU-Net}} 
 & D & $\mathbf{0.964 \pm 0.012}$ & $\mathbf{0.712 \pm 0.036}$ & $0.830 \pm 0.020$ & $\mathbf{0.765 \pm 0.023}$ \\
 & P & $\mathbf{0.962 \pm 0.016}$ & $0.764 \pm 0.019$ & $0.807 \pm 0.022$ & $0.706 \pm 0.027$ \\
 & R & $\mathbf{0.968 \pm 0.009}$ & $\mathbf{0.714 \pm 0.051}$ & $\mathbf{0.916 \pm 0.043}$ & $0.904 \pm 0.021$ \\
\bottomrule
\end{tabular}
} 
\end{table}

\begin{table}[t]
\centering
\caption{Test set performance. All metrics are macro-averaged on a per-case basis. Abbreviations: D (Dice), P (Precision), R (Recall), L (Liver), T (CRLM Tumor), E2E (End-to-End).}
\label{tab:test_results}
\setlength{\tabcolsep}{.6mm}
\resizebox{\columnwidth}{!}{
\begin{tabular}{l c cccccc}
\toprule
 & & \multicolumn{3}{c}{\textbf{Pipelined Tasks}} & \multicolumn{2}{c}{\textbf{Cascaded Tasks}} & \multicolumn{1}{c}{\textbf{E2E Task}} \\
\cmidrule(r){3-5} \cmidrule(l){6-7} \cmidrule(l){8-8}
\textbf{Model} & & 
\makecell{\textbf{Liver} \\ \textbf{(CT)}} & 
\makecell{\textbf{CRLM} \\ \textbf{(L+CT)}} & 
\makecell{\textbf{FLR} \\ \textbf{(L+T+CT)}} &
\makecell{\textbf{CRLM} \\ \textbf{(L+CT)}} &
\makecell{\textbf{FLR} \\ \textbf{(L+T+CT)}} &
\makecell{\textbf{FLR} \\ \textbf{(CT)}} \\
\midrule
\multirow{3}{*}{\textbf{nnU-Net}} 
 & D & $0.944$ & $0.588$ & $0.815$ & $0.525$ & $\textbf{0.767}$ & $0.757$\\
 & P & $0.944$ & $0.605$ & $\textbf{0.815}$ & $0.631$ & $\textbf{0.768}$ & $0.684$\\
 & R & $0.945$ & $\textbf{0.641}$ & $0.883$ & $0.520$ & $0.831$ & $\textbf{0.915}$\\
\midrule
\multirow{3}{*}{\makecell{\textbf{Swin-} \\ \textbf{UNETR}}} 
 & D & $0.969$ & $0.445$ & $0.742$ & $0.379$ & $0.694$ & $0.754$\\
 & P & $0.971$ & $0.635$ & $0.727$ & $0.549$ & $0.639$ & $\textbf{0.738}$\\
 & R & $0.967$ & $0.388$ & $0.855$ & $0.332$ & $0.838$ & $0.855$\\
\midrule
\multirow{3}{*}{\textbf{STU-Net}} 
 & D & $\textbf{0.973}$ & $\textbf{0.620}$ & $\textbf{0.834}$ & $\textbf{0.594}$ & $0.764$ & $\textbf{0.762}$\\
 & P & $\textbf{0.973}$ & $\textbf{0.686}$ & $0.811$ & $\textbf{0.653}$ & $0.732$ & $0.728$\\
 & R & $\textbf{0.974}$ & $0.623$ & $\textbf{0.919}$ & $\textbf{0.593}$ & $\textbf{0.856}$ & $0.896$\\
\bottomrule
\end{tabular}
} 
\end{table}

\section{Results and Discussion}
We present 5-fold CV results to demonstrate model stability and held-out test set results to establish final baseline performance (Tables~\ref{tab:cross_val_results} and \ref{tab:test_results}, respectively).

\noindent
\textbf{4.1 Baseline Performance:} The CV results (Table~\ref{tab:cross_val_results}) demonstrate low standard deviations, confirming training stability. STU-Net achieved the highest mean Dice across most tasks, notably for Liver ($0.964$), CRLM ($0.712$), and the E2E FLR task ($0.765$). nnU-Net secured the top performance for the Stage 3 FLR task ($0.834$), slightly ahead of STU-Net ($0.830$). On the test set (Table~\ref{tab:test_results}), the cascaded approach marginally outperformed E2E. The cascaded \textbf{nnU-Net} achieved the top FLR Dice (\textbf{0.767}), narrowly beating the cascaded STU-Net ($0.764$) and E2E STU-Net ($0.762$). Swin-UNETR was not competitive, particularly on CRLM segmentation.

\noindent
\textbf{4.2 Model Performance Discussion:} Our results highlight three key discussion points. First, \textbf{robustness to cascaded errors} is critical. As seen in Table~\ref{tab:test_results}, when moving from ground-truth inputs (``Pipelined'') to predicted inputs (``Cascaded'') for CRLM segmentation, STU-Net's Dice score dropped by only 0.026 ($0.620 \to 0.594$). In contrast, nnU-Net's performance fell by 0.063 ($0.588 \to 0.525$). This suggests that STU-Net's CRLM segmentation model is significantly more robust to the noisy, imperfect liver masks generated by the first stage of the cascade. Second, the \textbf{superior performance of STU-Net is likely attributable to pretraining}. We initialized STU-Net with its official pretrained weights, as recommended by the authors \cite{stunet}. This pretraining on large-scale medical datasets provides a clear advantage over nnU-Net and Swin-UNETR, which were trained from scratch. This performance hierarchy (STU-Net $>$ nnU-Net $>$ Swin-UNETR) is consistent with results reported on other large-scale segmentation benchmarks \cite{stunet}. Third, the \textbf{poor performance of Swin-UNETR} is likely due to its nature as a data-hungry transformer architecture. This is particularly evident for the CRLM task, whose complex metastatic patterns and small satellite lesions (excluded from labels for specificity) are challenging.

\section{Conclusion and Future Directions}
In this work, we present the first fully-manual, high-fidelity segmentation dataset for CRLM and FLR analysis. We establish the first segmentation baselines for FLR prediction, demonstrating that a cascaded approach is slightly superior to an E2E one. Pretrained models like STU-Net show high performance and greater robustness to cascaded errors, though the highly-optimized nnU-Net framework remains a top contender, achieving the best final FLR result. Future work will focus on improving the cascade's weakest link—CRLM segmentation—and exploring multi-task learning frameworks. We also acknowledge that our FLR ground truth is derived from a limited number of physicians. A key future direction is to incorporate multi-institutional data or consensus segmentations from multiple surgeons to better model surgical variability, though this presents a trade-off against validation on the final resected specimen.


\section{Acknowledgments}
The authors report no conflicts of interest.
\section{Compliance with Ethical Standards}
This research study was conducted retrospectively using human subject data made available in open access by CRLM-CT-Seg \cite{CRLMData}. Ethical approval was not required as confirmed by the license attached with the open access data. 


\bibliographystyle{IEEEbib}
\bibliography{strings,refs}

@article{bray2024global,
  title={Global cancer statistics 2022: GLOBOCAN estimates of incidence and mortality worldwide for 36 cancers in 185 countries},
  author={Bray, Freddie and Laversanne, Mathieu and Sung, Hyuna and Ferlay, Jacques and Siegel, Rebecca L and Soerjomataram, Isabelle and Jemal, Ahmedin},
  journal={CA: a cancer journal for clinicians},
  volume={74},
  number={3},
  pages={229--263},
  year={2024},
  publisher={Wiley Online Library}
}

@article{kron2024new,
  title={New trends in surgery for colorectal liver metastasis},
  author={Kron, Philipp and Lodge, Peter},
  journal={Annals of Gastroenterological Surgery},
  volume={8},
  number={4},
  pages={553--565},
  year={2024},
  publisher={Wiley Online Library}
}

@article{helling2014cause,
  title={Cause of death from liver metastases in colorectal cancer},
  author={Helling, Thomas S and Martin, Magdeline},
  journal={Annals of surgical oncology},
  volume={21},
  number={2},
  pages={501--506},
  year={2014},
  publisher={Springer}
}

@article{chow2019colorectal,
  title={Colorectal liver metastases: An update on multidisciplinary approach},
  author={Chow, Felix Che-Lok and Chok, Kenneth Siu-Ho},
  journal={World journal of hepatology},
  volume={11},
  number={2},
  pages={150},
  year={2019}
}

@article{calderon2023pushing,
  title={Pushing the limits of surgical resection in colorectal liver metastasis: how far can we go?},
  author={Calderon Novoa, Francisco and Ardiles, Victoria and de Santibanes, Eduardo and Pekolj, Juan and Goransky, Jeremias and Mazza, Oscar and S{\'a}nchez Claria, Rodrigo and de Santibanes, Martin},
  journal={Cancers},
  volume={15},
  number={7},
  pages={2113},
  year={2023},
  publisher={MDPI}
}

@article{veerankutty2021artificial,
  title={Artificial Intelligence in hepatology, liver surgery and transplantation: Emerging applications and frontiers of research},
  author={Veerankutty, Fadl H and Jayan, Govind and Yadav, Manish Kumar and Manoj, Krishnan Sarojam and Yadav, Abhishek and Nair, Sindhu Radha Sadasivan and Shabeerali, TU and Yeldho, Varghese and Sasidharan, Madhu and Rather, Shiraz Ahmad},
  journal={World Journal of Hepatology},
  volume={13},
  number={12},
  pages={1977},
  year={2021}
}

@article{kauffmann2014post,
  title={Post-hepatectomy liver failure},
  author={Kauffmann, Rondi and Fong, Yuman},
  journal={Hepatobiliary surgery and nutrition},
  volume={3},
  number={5},
  pages={238},
  year={2014}
}

@article{salavracos2024contribution,
  title={Contribution of 3D virtual modeling in locating hepatic metastases, particularly “vanishing tumors”: a pilot study},
  author={Salavracos, Mike and Danse, Etienne and Michoux, Nicolas and de Hemptinne, Alexandre and De Poortere, Tancr{\`e}de and Coubeau, Laurent},
  journal={Artificial Intelligence Surgery},
  volume={4},
  number={4},
  pages={331--347},
  year={2024},
  publisher={OAE Publishing Inc.}
}

@article{rompianesi2022artificial,
  title={Artificial intelligence in the diagnosis and management of colorectal cancer liver metastases},
  author={Rompianesi, Gianluca and Pegoraro, Francesca and Ceresa, Carlo DL and Montalti, Roberto and Troisi, Roberto Ivan},
  journal={World Journal of Gastroenterology},
  volume={28},
  number={1},
  pages={108},
  year={2022}
}

@article{amygdalos2023novel,
  title={Novel machine learning algorithm can identify patients at risk of poor overall survival following curative resection for colorectal liver metastases},
  author={Amygdalos, Iakovos and M{\"u}ller-Franzes, Gustav and Bednarsch, Jan and Czigany, Zoltan and Ulmer, Tom Florian and Bruners, Philipp and Kuhl, Christiane and Neumann, Ulf Peter and Truhn, Daniel and Lang, Sven Arke},
  journal={Journal of Hepato-Biliary-Pancreatic Sciences},
  volume={30},
  number={5},
  pages={602--614},
  year={2023},
  publisher={Wiley Online Library}
}

@article{mehrabi2018meta,
  title={Meta-analysis of the prognostic role of perioperative platelet count in posthepatectomy liver failure and mortality},
  author={Mehrabi, Arianeb and Golriz, Mohammad and Khajeh, Elias and Ghamarnejad, Omid and Probst, Pascal and Fonouni, Hamidreza and Mohammadi, Sara and Weiss, Karl Heinz and B{\"u}chler, Markus W},
  journal={Journal of British Surgery},
  volume={105},
  number={10},
  pages={1254--1261},
  year={2018},
  publisher={Oxford University Press}
}

@ARTICLE{itksnap,
  author = {Paul A. Yushkevich and Joseph Piven and Cody Hazlett, Heather and
    Gimpel Smith, Rachel and Sean Ho and James C. Gee and Guido Gerig},
  title = {User-Guided {3D} Active Contour Segmentation of
    Anatomical Structures: Significantly Improved Efficiency and Reliability},
  journal = {Neuroimage},
  year = {2006},
  volume = {31},
  number = {3},
  pages = {1116--1128},
}

@article{nnUnet,
  title={nnU-Net: a self-configuring method for deep learning-based biomedical image segmentation},
  author={Isensee, Fabian and Jaeger, Paul F and Kohl, Simon AA and Petersen, Jens and Maier-Hein, Klaus H},
  journal={Nature methods},
  volume={18},
  number={2},
  pages={203--211},
  year={2021},
  publisher={Nature Publishing Group}
}

@inproceedings{swinUnetr,
  title={Swin unetr: Swin transformers for semantic segmentation of brain tumors in mri images},
  author={Hatamizadeh, Ali and Nath, Vishwesh and Tang, Yucheng and Yang, Dong and Roth, Holger R and Xu, Daguang},
  booktitle={International MICCAI brainlesion workshop},
  pages={272--284},
  year={2021},
  organization={Springer}
}

@article{stunet,
  title={Stu-net: Scalable and transferable medical image segmentation models empowered by large-scale supervised pre-training},
  author={Huang, Ziyan and Wang, Haoyu and Deng, Zhongying and Ye, Jin and Su, Yanzhou and Sun, Hui and He, Junjun and Gu, Yun and Gu, Lixu and Zhang, Shaoting and others},
  journal={arXiv preprint arXiv:2304.06716},
  year={2023}
}

@article{CRLMData,
  title={Preoperative CT and survival data for patients undergoing resection of colorectal liver metastases},
  author={Simpson, Amber L and Peoples, Jacob and Creasy, John M and Fichtinger, Gabor and Gangai, Natalie and Keshavamurthy, Krishna N and Lasso, Andras and Shia, Jinru and D’Angelica, Michael I and Do, Richard KG},
  journal={Scientific Data},
  volume={11},
  number={1},
  pages={172},
  year={2024},
  publisher={Nature Publishing Group UK London}
}

\end{document}